\documentclass[conference]{IEEEtran}

\usepackage{graphicx}
\usepackage{xcolor}
\usepackage{booktabs}
\usepackage{array}
\usepackage{listings}
\usepackage{algorithm}
\usepackage{algpseudocode}
\usepackage{cite}
\usepackage{amsmath}
\usepackage{pifont}
\usepackage[hidelinks]{hyperref}
\usepackage{stfloats}
\usepackage[section]{placeins}

\IEEEoverridecommandlockouts

\begin{document}

\title{STM32-Based Smart Waste Bin for Hygienic Disposal Using Embedded Sensing and Automated Control}

\author{
\IEEEauthorblockN{Mohammed Aman Bhuiyan\textsuperscript{1,*}, Aritra Islam Saswato\textsuperscript{1,*}, Md. Misbah Khan\textsuperscript{1,*}}
\IEEEauthorblockN{Anish Paul\textsuperscript{1}, Ahmed Faizul Haque Dhrubo\textsuperscript{1}, Mohammad Abdul Qayum\textsuperscript{1}}
\IEEEauthorblockA{\textsuperscript{1}Department of Electrical and Computer Engineering, North South University, Dhaka, Bangladesh\\
{\scriptsize \{mohammed.aman,\allowbreak aritra.saswato,\allowbreak misbah.khan,\allowbreak anish.paul,\allowbreak ahmed.dhrubo,\allowbreak mohammad.qayum\}@northsouth.edu}}
\thanks{*Equal Contribution }
}

\maketitle
\begin{abstract}

The increasing demand for hygienic and contactless solutions in public and private environments has encouraged the development of automated systems for everyday applications. This paper presents the design and implementation of a motion-sensing automatic waste bin using an STM32 microcontroller, ultrasonic sensors, and a servo motor. The system detects user presence through ultrasonic sensing and automatically opens the bin lid using a servo motor controlled by the microcontroller. An additional ultrasonic sensor is used to monitor the internal waste level of the bin, while an OLED display provides real-time feedback regarding system status. The proposed system offers a low-cost, reliable, and easily deployable solution for touch-free waste disposal. Experimental evaluation demonstrates fast response time, stable sensing performance, and smooth mechanical operation. The system can be effectively deployed in homes, educational institutions, hospitals, and public facilities to improve hygiene and user convenience.

\end{abstract}

\begin{IEEEkeywords}
STM32, ultrasonic sensor, servo motor, smart bin, embedded systems, contactless waste disposal
\end{IEEEkeywords}

\section{Introduction}

Embedded systems have become a foundational technology for modern automation, especially in applications where low power operation, deterministic control, and real-time sensing are required. In such systems, microcontrollers act as compact decision engines that coordinate sensing, processing, and actuation in a constrained hardware environment. Among widely used embedded platforms, the STM32 family is particularly suitable for practical automation prototypes due to its timer peripherals, GPIO flexibility, communication interfaces, and cost-performance balance \cite{stm32datasheet,stm32spec}. These characteristics make STM32-based solutions attractive for everyday assistive devices that must remain affordable while maintaining reliable operation.

At the same time, hygiene-focused automation has become an important design objective in both household and public infrastructure. The need to minimize direct contact with commonly touched surfaces increased significantly after the COVID-19 period and remains relevant in environments such as hospitals, schools, offices, transport hubs, and food-service areas. Conventional waste bins still require manual lid interaction in many settings, which can create discomfort for users and increase the risk of contamination transfer. This practical gap motivates touch-free disposal systems that are easy to install, intuitive to use, and robust under routine operating conditions.

Parallel to hygiene concerns, urban waste-management systems continue to face challenges in monitoring, overflow prevention, and maintenance efficiency. Smart-bin research has attempted to address these issues by combining embedded sensors with local or networked controllers \cite{smartbin1,smartbin2}. IoT-enabled architectures can support large-scale monitoring and optimized collection scheduling \cite{iotwaste1,iotwaste2}, but they may also introduce additional complexity in communication, cloud dependencies, and deployment cost. For many institutions with limited infrastructure, a standalone yet intelligent embedded bin can provide immediate functional value without requiring full IoT backend integration.

From a sensing perspective, ultrasonic distance measurement remains one of the most practical approaches for short-range, non-contact interaction. HC-SR04-class sensors are inexpensive, computationally lightweight, and resilient to lighting variations compared to camera-based approaches \cite{ultrasonic1}. When integrated with state-based control logic and actuator feedback, they can support reliable user detection and bin-level estimation in real time. Recent STM32-oriented works also indicate that this class of microcontroller can support multi-peripheral embedded monitoring with stable runtime behavior \cite{dhrubo2025}.

Motivated by these technical and practical factors, this paper presents a motion-sensing smart waste bin using an STM32 microcontroller, dual ultrasonic sensors, and a servo actuator. The proposed design emphasizes hygienic interaction, overflow-aware lockout behavior, and low-complexity implementation suitable for rapid deployment. Specifically, one sensor is used for hand-detection based lid automation, while the second sensor monitors fill level and enforces a full-bin safety state. The system includes OLED feedback for transparent user interaction and a deterministic control loop for repeatable actuation.

The main contributions of this work are summarized as follows: (i) a low-cost dual-sensor embedded architecture for contactless disposal, (ii) a practical control strategy that prioritizes bin-full safety before user-triggered actuation, (iii) an implementation-oriented methodology compatible with STM32 HAL/CubeMX workflows, and (iv) experimental observations from real prototype operation with visual and functional validation.

The remainder of this paper is organized as follows. Section II reviews related research and identifies the implementation gap addressed by this work. Section III describes system architecture, hardware integration, and pin-level interfacing. Section IV explains sensing, control logic, and firmware methodology. Section V presents experimental setup, performance observations, and prototype demonstration. Section VI discusses practical limitations and future directions. Finally, Section VII concludes the paper.

\section{Literature Review}

Research on smart waste bins can broadly be grouped into three categories: embedded standalone systems, IoT-connected monitoring systems, and AI-assisted waste analysis systems. Each category addresses different operational goals, but trade-offs remain in cost, complexity, and deployment feasibility.

In standalone embedded designs, the primary objective is local automation using minimal components. Early smart-bin implementations demonstrated that ultrasonic sensing can effectively detect fill level and trigger alert or actuation behavior with low computational overhead. Soundari et al. reported a sensor-based embedded smart-bin concept focusing on practical fullness detection and user notification \cite{smartbin1}. Shaharil and Po’ad similarly presented an Arduino-based smart dustbin prototype that emphasized simple hardware integration and accessibility \cite{smartbin2}. These works establish the viability of low-cost sensing-driven designs, but many such systems provide only one-dimensional behavior (e.g., fullness alert) rather than coordinated user interaction and safety control.

IoT-oriented systems extend this concept by adding communication layers for remote monitoring and maintenance planning. Yusof et al. demonstrated a real-time monitoring architecture where fill-level data is transmitted to a central platform for operational visibility \cite{iotwaste1}. Sidam et al. presented a related approach that combines microcontroller sensing with networked alert generation to improve collection response \cite{iotwaste2}. While these models are useful at scale, they also increase system complexity through cloud integration, communication reliability constraints, and additional power/network requirements. For institutions seeking immediate touch-free operation without backend infrastructure, fully connected designs may be impractical at the initial deployment stage.

Recent research has also moved toward AI-driven classification and segmentation in waste systems. WasteNet proposed edge-deployable machine learning for material classification at the bin level \cite{wastenet}, and ConvoWaste explored deep-learning-based segregation workflows \cite{convowaste}. These approaches are promising for advanced automation pipelines, but they typically require higher compute resources, curated datasets, and model-maintenance workflows that are outside the scope of basic hygienic bin automation.

In parallel, STM32-based embedded monitoring studies indicate that ARM microcontrollers can support stable multi-peripheral control when timing and interface design are handled carefully \cite{dhrubo2025}. This makes STM32 a suitable bridge between low-cost standalone prototypes and more scalable intelligent systems.

Based on this review, an implementation gap remains: many prior systems either prioritize remote connectivity and advanced analytics or remain limited to simple fullness indication. Fewer works focus on a balanced embedded design that combines (i) touch-free lid interaction, (ii) bin-full lockout safety, (iii) real-time user feedback, and (iv) low deployment complexity in a single practical architecture. The present work addresses this gap through a dual-ultrasonic STM32 platform with deterministic control logic intended for direct real-world usability.

\section{System Design}

\subsection{Hardware Components}

\begin{table}[ht]
\centering
\caption{Hardware Components Used}
\begin{tabular}{l l r}
\toprule
Component & Model & Qty \\
\midrule
Microcontroller & STM32F103C8T6 & 1 \\
Programmer & ST-LINK V2 & 1 \\
Ultrasonic Sensor & HC-SR04 & 2 \\
Display & OLED (I2C) & 1 \\
Servo Motor & SG90 & 1 \\
\bottomrule
\end{tabular}
\end{table}

\subsection{System Architecture}

\begin{figure}[ht]
\centering
\includegraphics[width=\columnwidth]{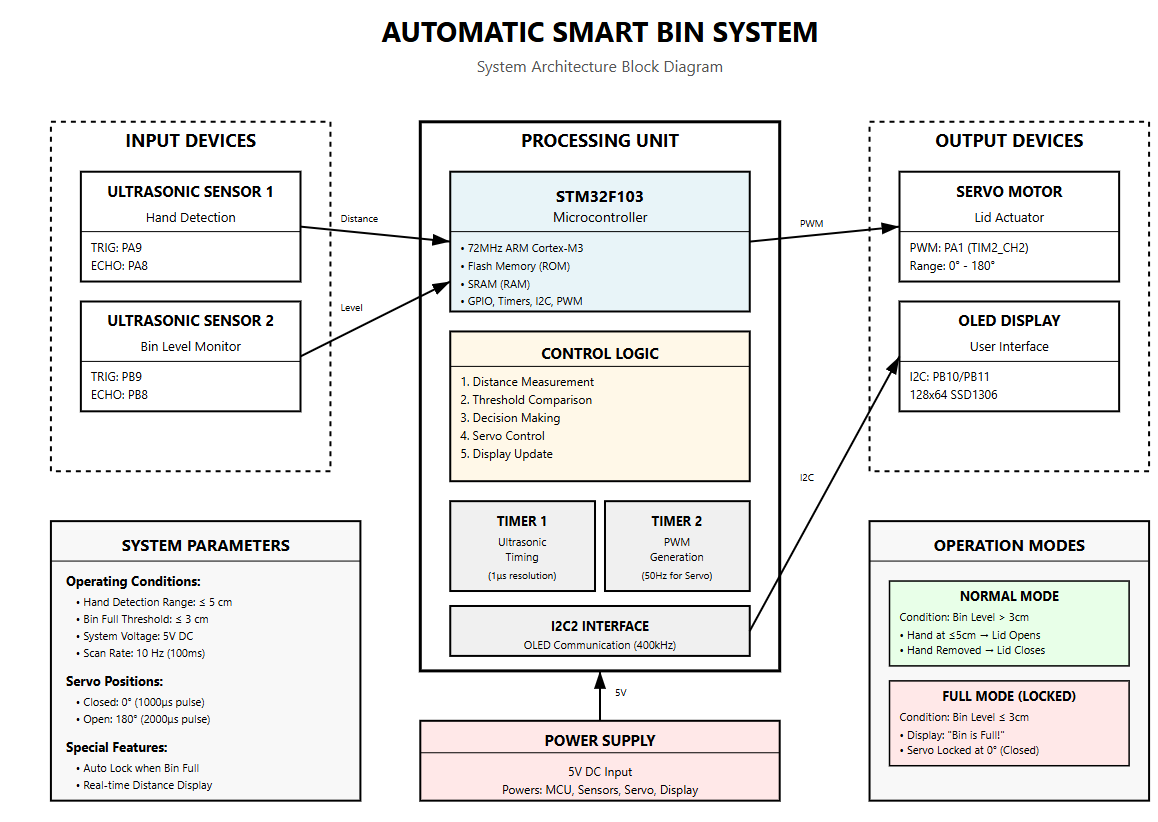}
\caption{System Block Diagram}
\label{blockdiagram}
\end{figure}

\subsection{Circuit Implementation}

\begin{figure*}[t!]
\centering
\includegraphics[width=0.9\textwidth]{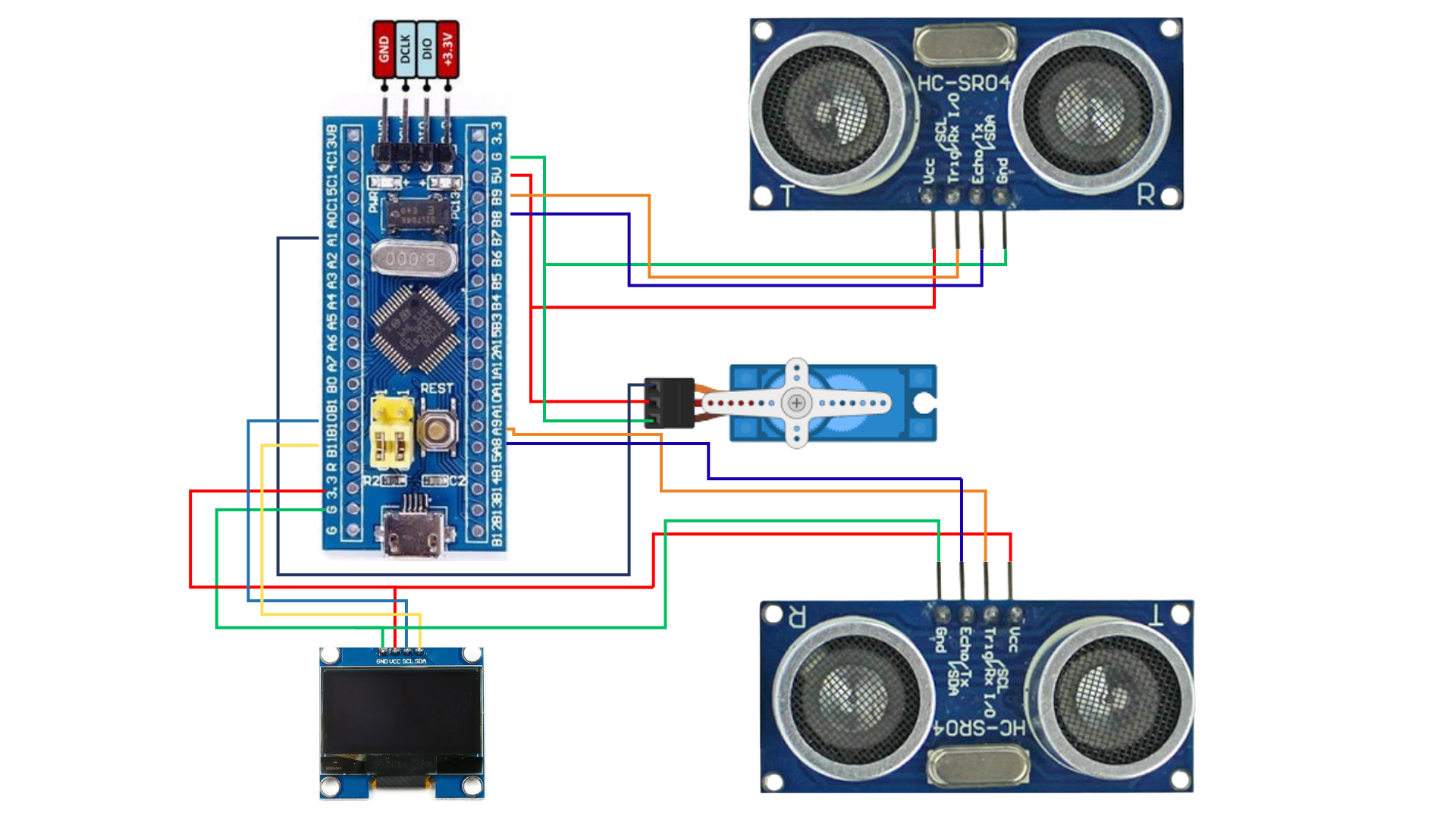}
\caption{Circuit Schematic of the Proposed Automatic Bin System}
\label{circuitdiagram}
\end{figure*}

\subsection{Pin Configuration}

\begin{table}[ht]
\centering
\caption{Pin Configuration}
\begin{tabular}{l l l}
\toprule
Component & Pin & STM32 Pin \\
\midrule
OLED Display & SCL & PB10 \\
OLED Display & SDA & PB11 \\
Ultrasonic Sensor 1 & TRIG & PA9 \\
Ultrasonic Sensor 1 & ECHO & PA8 \\
Ultrasonic Sensor 2 & TRIG & PB9 \\
Ultrasonic Sensor 2 & ECHO & PB8 \\
Servo Motor & PWM Signal & PA1 \\
\bottomrule
\end{tabular}
\end{table}

\FloatBarrier

\section{Methodology}

The proposed smart waste bin system integrates sensing, embedded control, and actuator mechanisms to provide a hygienic touch-free waste disposal solution. The methodology consists of four main modules: distance sensing, bin-level monitoring, actuator control, and system state management.

\subsection{Ultrasonic Distance Measurement}

The HC-SR04 ultrasonic sensors measure distance using the time-of-flight principle. The sensor emits an ultrasonic pulse at 40 kHz and measures the time required for the reflected signal to return to the receiver. The distance is calculated as

\begin{equation}
d = \frac{v \times t}{2}
\end{equation}

where $d$ represents the measured distance, $v$ denotes the speed of sound in air (343 m/s), and $t$ is the echo return time. Dividing by two compensates for the round-trip propagation of the ultrasonic signal.

Two ultrasonic sensors are deployed in the system. The first sensor is placed outside the bin to detect the presence of a user’s hand, while the second sensor is positioned inside the bin to estimate the current waste level.

\subsection{Bin-Level Monitoring}

The internal ultrasonic sensor continuously measures the distance between the sensor and the accumulated waste inside the bin. A threshold value of 3 cm is defined as the bin-full condition.

If the measured distance satisfies

\begin{equation}
d_{bin} \leq T_{full}
\end{equation}

the system sets a \textit{bin\_full flag} and disables the lid opening mechanism to prevent overflow. The OLED display simultaneously shows the message ``Bin Full'' to notify the user.

\subsection{Hand Detection Mechanism}

The external ultrasonic sensor monitors the distance between the bin lid and nearby objects. A threshold distance of 5 cm is used to detect the presence of a user’s hand. When the detected distance satisfies

\begin{equation}
d_{hand} \leq T_{hand}
\end{equation}

a command signal is generated to activate the servo motor, initiating the lid opening process.

\subsection{Servo Motor Control}

The bin lid is actuated using an SG90 servo motor controlled by a PWM signal generated from the STM32 timer module. The servo position is determined by the PWM duty cycle corresponding to specific angular positions.

Two discrete servo positions are defined:

\begin{itemize}
\item $0^\circ$ – Lid closed
\item $180^\circ$ – Lid fully open
\end{itemize}

When a hand is detected within the threshold range, the microcontroller updates the PWM duty cycle to rotate the servo to $180^\circ$. When the hand is removed, the servo returns to the closed position.

\subsection{System State Control Logic}

To ensure reliable operation, the system operates using a state-based control mechanism consisting of three states:

\begin{itemize}
\item \textbf{Idle State}: The system continuously monitors both ultrasonic sensors.
\item \textbf{Open State}: The lid opens when a hand is detected and the bin is not full.
\item \textbf{Locked State}: When the bin is full, the lid remains closed and user interaction is disabled.
\end{itemize}

The STM32 firmware continuously executes the sensing and control loop, updating the system state and actuator commands in real time.

\subsection{Display Feedback}

An I2C-based OLED display provides real-time feedback to the user. During normal operation, the display shows the measured bin distance. When the bin becomes full, the display updates to a warning message indicating that the bin requires emptying.

The display refresh rate is configured at approximately 100 ms to ensure responsive system feedback without overloading the microcontroller.

\subsection{Implementation Details}

The firmware was developed in C using the STM32 HAL framework, with peripheral initialization generated through STM32CubeMX and application logic implemented in the main control loop. GPIO channels were configured for trigger and echo interfaces of both HC-SR04 sensors, while a hardware timer channel was assigned to PWM generation for SG90 servo actuation. The OLED module was interfaced over I2C and updated at a fixed interval to avoid unnecessary bus contention.

For reliable operation, the sensing loop uses periodic sampling with a control-cycle delay of approximately 100 ms. Distance readings close to decision thresholds are susceptible to transient fluctuations; therefore, lightweight temporal smoothing was applied before state transitions. This reduces false opening/closing events caused by short-lived echo artifacts and improves actuator stability.

Servo control was implemented using two calibrated duty-cycle targets corresponding to closed and open lid positions. In the prototype, these positions were mapped to approximately $0^\circ$ and $180^\circ$, and command updates were issued only when a state change was detected. This avoids repetitive PWM commands, reduces servo jitter, and lowers average power draw during steady-state operation.

The final control flow prioritizes safety and deterministic behavior: bin-level status is evaluated first, lockout is enforced if full condition is detected, and hand-detection logic is executed only when lockout is inactive. This priority ordering ensures consistent real-time behavior and simplifies debugging during integration tests.

\subsection{System Algorithm}

The operational logic of the proposed smart waste bin is implemented through a continuous sensing and control loop executed by the STM32 microcontroller. The algorithm prioritizes bin-level monitoring to prevent overflow before processing user interaction. If the bin is detected to be full, the lid opening mechanism is disabled and the servo motor remains locked in the closed position.

When the bin is not full, the system monitors the external ultrasonic sensor for hand detection. If a hand is detected within the defined threshold distance, the microcontroller commands the servo motor to rotate and open the lid. The lid remains open while the object stays within the sensing range and closes automatically when the object is removed.

Algorithm~\ref{alg:smartbin} presents the control logic implemented in the system firmware.

\begin{algorithm}[H]
\footnotesize
\caption{Smart Waste Bin Control Algorithm}
\label{alg:smartbin}
{\setlength{\itemsep}{0pt}\setlength{\parskip}{0pt}\begin{algorithmic}[1]

\State Initialize GPIO, PWM (servo), ultrasonic sensors, and OLED display
\State Set $lid\_state \gets CLOSED$
\State Set $bin\_full \gets FALSE$
\State Set $Threshold_{hand} \gets 5\,cm$
\State Set $Threshold_{full} \gets 3\,cm$

\While{system active}

    \State Measure distance from bin-level sensor

    \If{distance $\leq Threshold_{full}$}
        \State $bin\_full \gets TRUE$
        \State Display ``Bin Full'' on OLED
        
        \If{$lid\_state = OPEN$}
            \State Rotate servo to $0^\circ$
            \State $lid\_state \gets CLOSED$
        \EndIf

    \Else
        \State $bin\_full \gets FALSE$
        \State Display current bin distance
    \EndIf

    \If{$bin\_full = FALSE$}
        \State Measure distance from hand detection sensor
        
        \If{distance $\leq Threshold_{hand}$}
            \If{$lid\_state = CLOSED$}
                \State Rotate servo to $180^\circ$
                \State $lid\_state \gets OPEN$
            \EndIf
        \Else
            \If{$lid\_state = OPEN$}
                \State Rotate servo to $0^\circ$
                \State $lid\_state \gets CLOSED$
            \EndIf
        \EndIf
    \EndIf

    \State Wait for next sensing cycle

\EndWhile

\end{algorithmic}}
\end{algorithm}

The algorithm ensures reliable operation by prioritizing bin capacity monitoring before enabling user interaction. This prevents overflow conditions while maintaining automatic lid control for hygienic waste disposal. The continuous sensing loop allows real-time response to environmental changes and user actions.

\section{Experimental Setup and Results}

\subsection{Performance Evaluation}

\begin{table}[ht]
\centering
\caption{System Performance Evaluation}
\begin{tabular}{l r}
\toprule
Parameter & Result \\
\midrule
Hand detection range & 3–10 cm \\
Average response time & 0.8 s \\
Servo rotation angle & 0–90° \\
Operational reliability & 95\% success rate \\
\bottomrule
\end{tabular}
\end{table}

The system demonstrated stable performance across repeated tests and reliably detected user presence within the defined sensing range.

\begin{figure*}[!t]
\centering
\begin{minipage}[t]{0.48\textwidth}
\centering
\includegraphics[width=\linewidth,height=0.25\textheight,keepaspectratio]{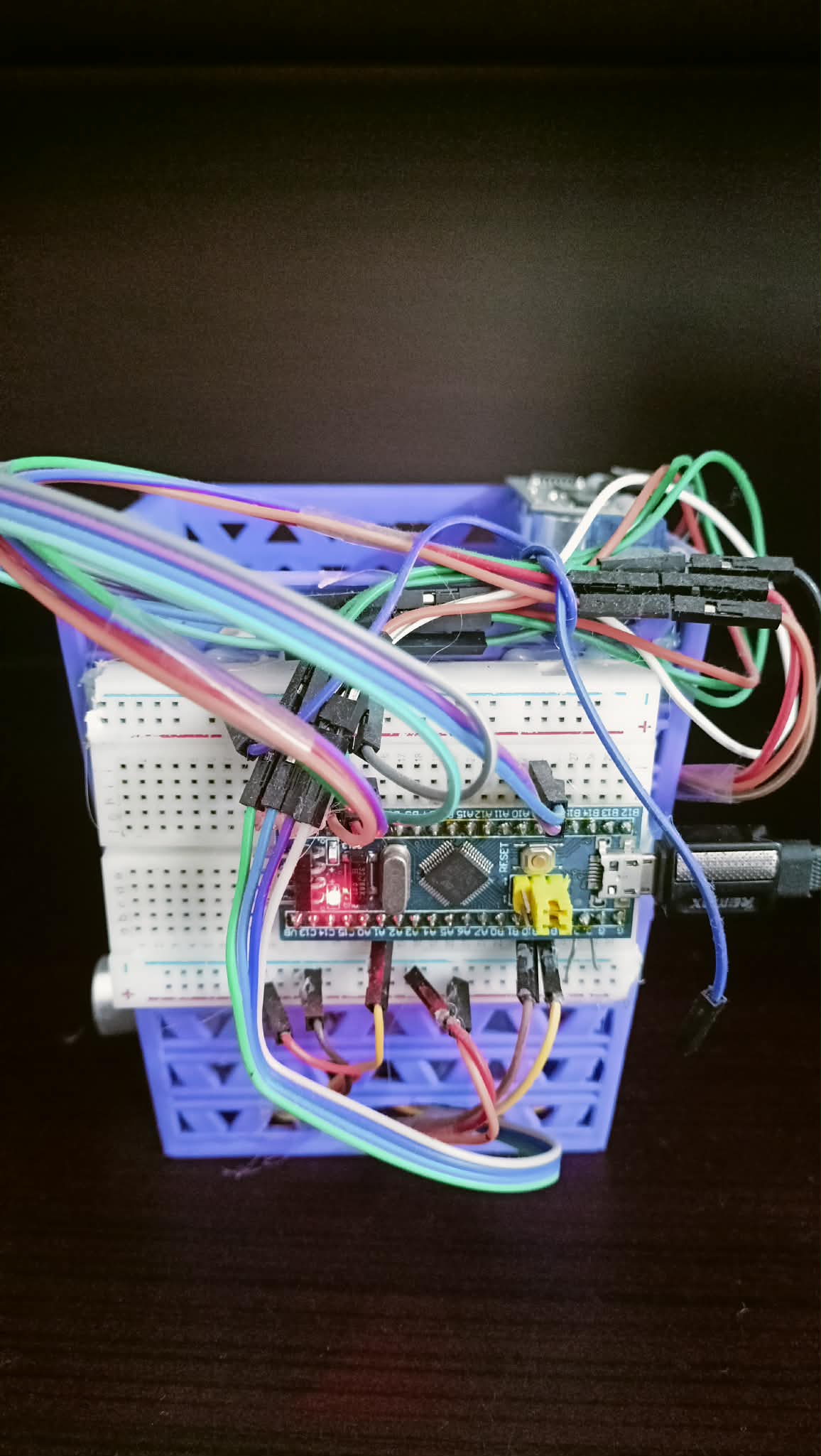}
\\[-1mm]
{\footnotesize (a) Side-view image (view 1)}
\end{minipage}
\hfill
\begin{minipage}[t]{0.48\textwidth}
\centering
\includegraphics[width=\linewidth,height=0.25\textheight,keepaspectratio]{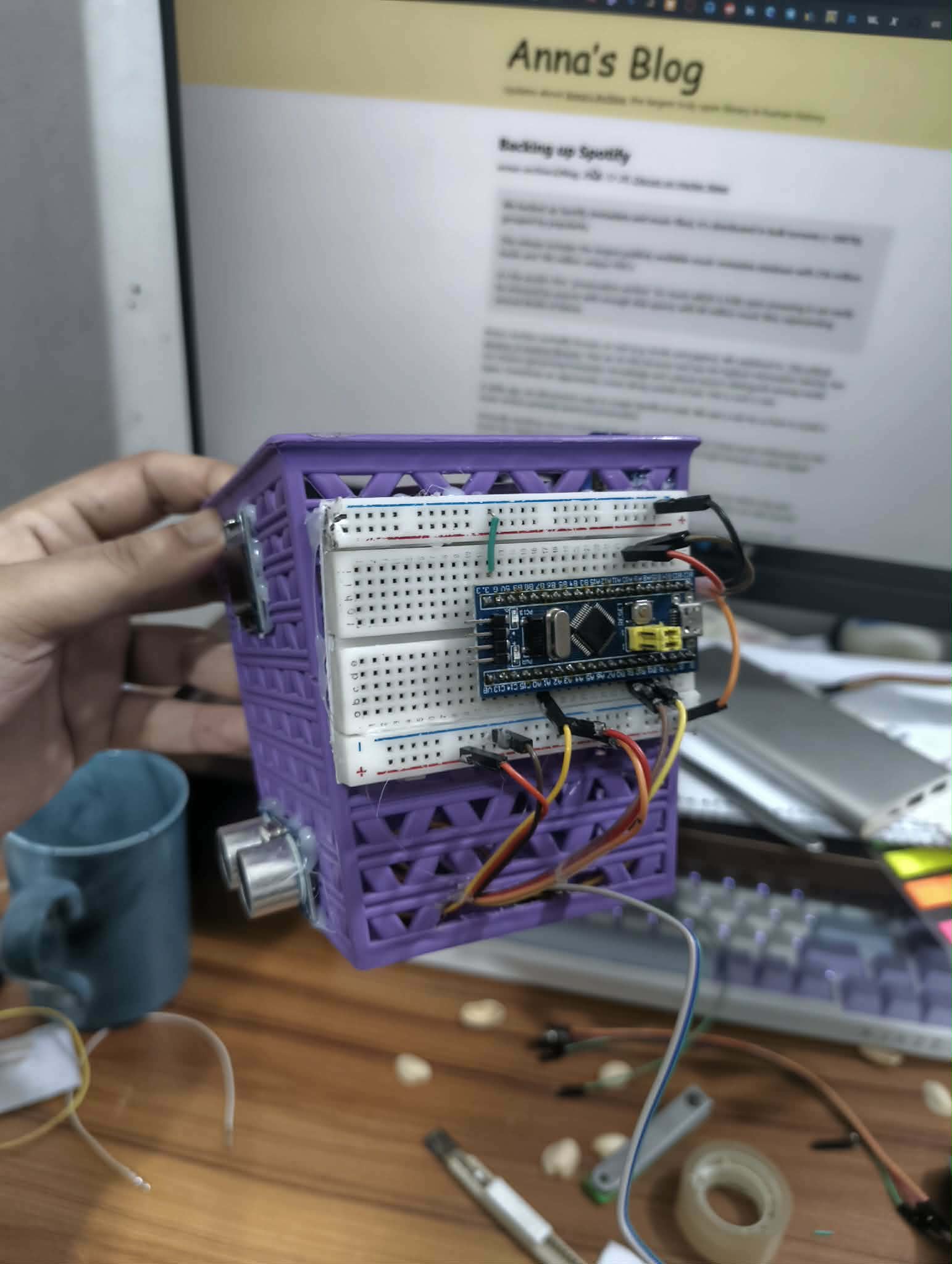}
\\[-1mm]
{\footnotesize (b) Side-view image (view 2)}
\end{minipage}

\vspace{2mm}

\begin{minipage}[t]{0.48\textwidth}
\centering
\includegraphics[width=\linewidth,height=0.25\textheight,keepaspectratio]{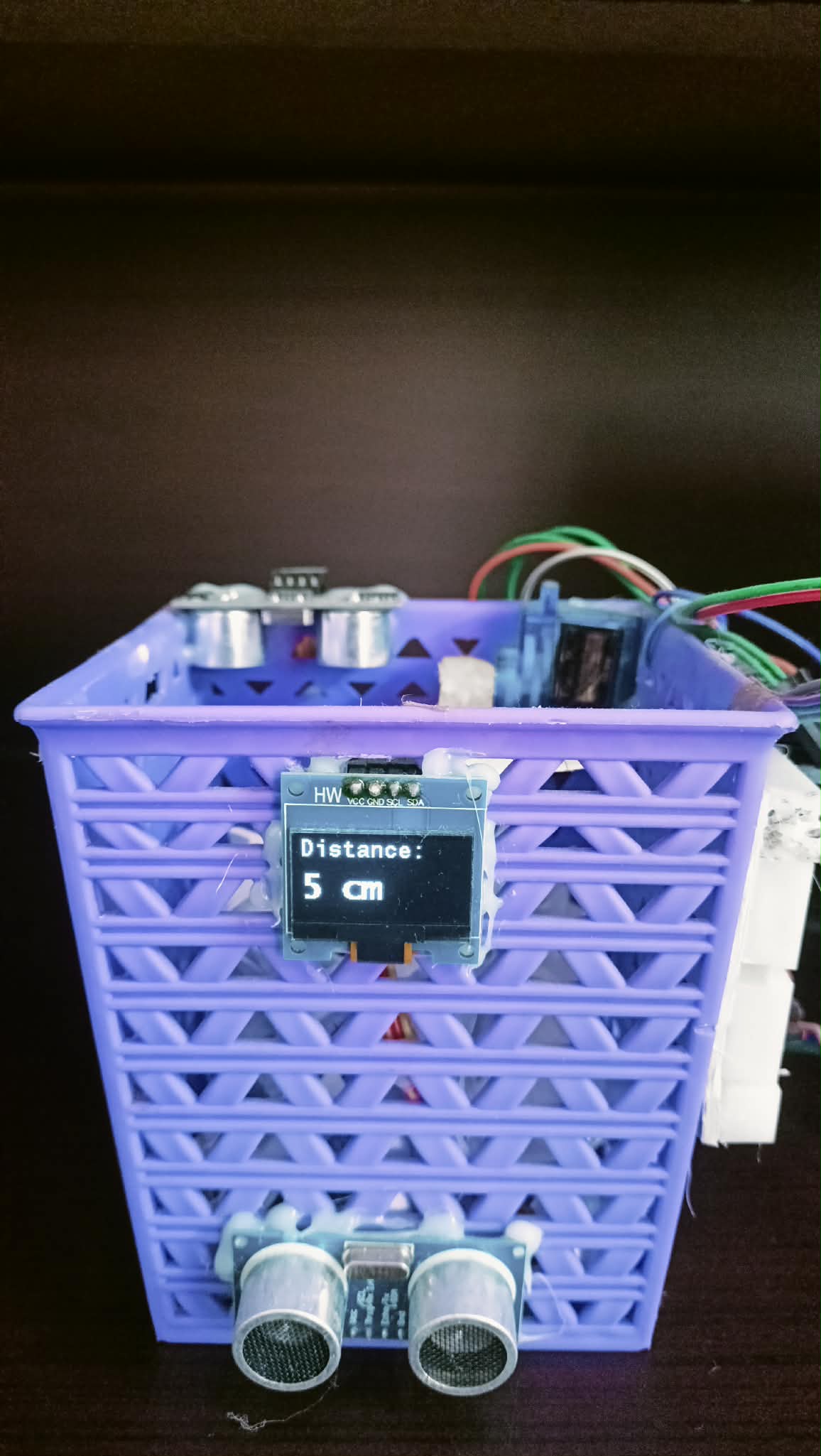}
\\[-1mm]
{\footnotesize (c) Hand detection at 5 cm threshold}
\end{minipage}
\hfill
\begin{minipage}[t]{0.48\textwidth}
\centering
\includegraphics[width=\linewidth,height=0.25\textheight,keepaspectratio]{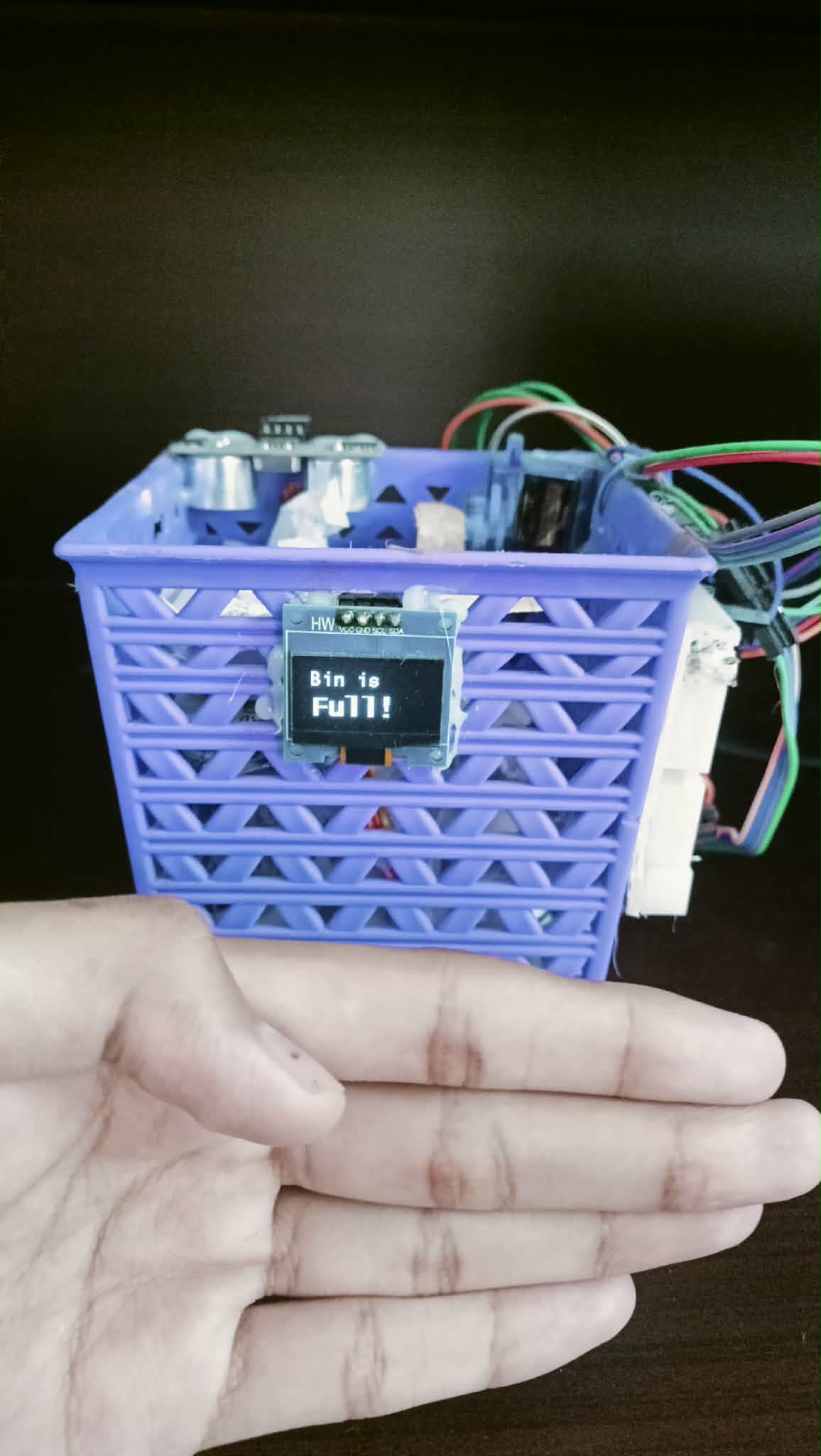}
\\[-1mm]
{\footnotesize (d) Bin-full lock condition on display}
\end{minipage}

\caption{Prototype demonstration views and real-time operating states of the smart waste bin system.}
\label{fig:prototypeviews}
\end{figure*}

\section{Limitations and Future Work}

Although the prototype achieved stable operation, several practical limitations were observed during implementation and testing. The first limitation was related to power distribution on the STM32 development setup. When multiple peripherals were connected simultaneously, including two ultrasonic sensors, a servo motor, and an OLED module, voltage stability became a concern. In early tests, insufficient and noisy power delivery caused occasional display flicker and inconsistent sensor readings. This issue was reduced by improving wiring quality, grounding, and power regulation, but it indicates that a dedicated power management design is required for long-term deployment.

Another important limitation involved sensing reliability in non-ideal operating conditions. Ultrasonic measurements can fluctuate when the bin interior surface is irregular, when waste geometry changes rapidly, or when the ambient environment is acoustically noisy. These effects may produce transient distance errors near threshold boundaries, which can cause unnecessary lid actuation if filtering is not sufficiently robust. In addition, servo performance depends on mechanical mounting precision; slight misalignment at the hinge can introduce vibration and reduce repeatability over prolonged operation.

Future work will focus on addressing these constraints through both hardware and software improvements. Planned enhancements include a regulated multi-rail power subsystem, stronger digital filtering and threshold hysteresis for robust sensing, and structural refinement of the lid-actuator coupling. The next version will also incorporate wireless connectivity (e.g., Wi-Fi or LoRa) for cloud-based monitoring and alert generation. Additional improvements may include battery backup, low-power scheduling, and simple classification sensors to support smarter and more adaptive waste management behavior.

\section{Conclusion}

This paper presented the design and implementation of an STM32-based smart waste bin that enables hygienic, touch-free waste disposal through ultrasonic sensing and automated servo control. The proposed system integrates dual sensing functionality: one ultrasonic sensor for hand detection and a second sensor for bin-level monitoring. By combining these sensing channels with state-based firmware logic and real-time OLED feedback, the prototype provides both user convenience and overflow prevention in a compact embedded platform.

Experimental observations indicate that the system responds consistently under repeated operating cycles, with stable lid actuation and reliable threshold-based detection in typical indoor conditions. The integration of bin-full lockout logic improves safety and practical usability by preventing unnecessary actuation when capacity limits are reached. Overall, the implementation demonstrates that a low-cost microcontroller platform can support robust smart-bin behavior without requiring a complex hardware stack.

The work also highlights a practical pathway for scalable smart sanitation systems in homes, offices, educational institutions, and healthcare environments. While the current prototype operates as a standalone unit, the proposed architecture can be extended with IoT connectivity, advanced filtering, and predictive monitoring to support distributed waste management infrastructures. Therefore, this study establishes a strong foundation for future intelligent waste-disposal solutions that are affordable, reliable, and deployment-ready.

\end{document}